\documentclass[letterpaper, 10 pt, conference]{ieeeconf}  

\IEEEoverridecommandlockouts                              

\usepackage[pdftex]{graphicx}
\usepackage{amsmath} 
\usepackage{amssymb}  
\usepackage{subfigure}
\usepackage{multirow}
\usepackage{array,booktabs}
\usepackage{diagbox}
\usepackage{balance}
\usepackage{mathtools}
\usepackage{subfigure}
\usepackage{verbatim}
\usepackage[table]{xcolor}
\makeatletter
\let\NAT@parse\undefined
\makeatother
\usepackage[numbers]{natbib}

\usepackage{irap_SIunits}
\usepackage{irap_acronyms}
\usepackage{irap_math}
\usepackage{irap_misc}

\usepackage{soul,color}

\usepackage{xcolor}
\newcommand{\bl}[1]{{\textcolor{black}{#1}}}

\DeclareMathOperator*{\argmin}{argmin}
	\DeclareMathOperator*{\argmax}{argmax}

\usepackage{caption}

\title{\LARGE \bf
Road is Enough! Extrinsic Calibration of Non-overlapping \\Stereo Camera and LiDAR using Road Information
}

\author{Jinyong Jeong${}^{1}$, Lucas Y. Cho${}^{1}$ and Ayoung Kim${}^{1*}$
\thanks{J. Jeong and A. Kim are with the Department of Civil and Environmental Engineering,
        KAIST, Daejeon, S. Korea \texttt{[jjy0923, dudgnsrj, ayoungk]@kaist.ac.kr}}%
\thanks{This work is supported through a grant from the KAIA grant funded
        by the MOLIT of Korea (19CTAP-C142170-02) and by [High-Definition Map Based Precise Vehicle Localization Using Cameras and LIDARs] project funded by Naver Labs Corporation.}%
}

\begin{document}

\maketitle
\thispagestyle{empty}
\pagestyle{empty}

\begin{abstract}

This paper presents a framework for the target-less extrinsic calibration of stereo cameras and \ac{LiDAR} sensors with a non-overlapping \ac{FOV}. \bl{In order to} solve the extrinsic calibrations problem under such challenging configuration, the proposed solution exploits road markings as static and robust features \bl{among the} various dynamic objects \bl{that are present} in urban environment. First, \bl{this study utilizes road markings that are commonly captured by the two sensor modalities to select informative images for estimating the extrinsic parameters.} \bl{In order to} accomplish stable optimization, multiple cost functions \bl{are defined,} including \ac{NID}, edge alignment and, plane fitting cost. \bl{Therefore} a smooth cost curve is formed for global optimization to prevent convergence to the local optimal point. We further evaluate each cost function by examining parameter sensitivity near the optimal point. Another key characteristic of extrinsic calibration, repeatability, is analyzed by conducting the proposed method multiple times with varying randomly perturbed initial points.

\end{abstract}


\section{Introduction} \label{introduction}

Autonomous vehicles are often equipped with \bl{several} multi-modal sensors for environmental perception. \bl{Typical perceptual sensors include} range sensors which measure the distance to objects using signals (e.g., light, radio waves, and sound waves) and vision sensors which project 3D information onto a 2D image frame. \bl{If multiple sensor types are equipped,} obtaining relative coordinate transformations via extrinsic calibration becomes an essential pre-processing module. \bl{According to} literature, accurate extrinsic calibration enhances the performance of data fusion between multiple sensors and the complementary exploitation of other modalities. For example,
\citeauthor{shin2018direct}~\cite{shin2018direct} utilized depth information from
\ac{LiDAR} to calculate the photometric error of two images.
\citeauthor{zhang2015visual}~\cite{zhang2015visual} implemented a fast odometry
method \bl{featuring} low drift by fusing visual odometry and \ac{LiDAR} odometry
methods. \citeauthor{han2017road}~\cite{han2017road} conducted a study to estimate road area using LiDAR data and camera images using a conditional random field framework.


The most common extrinsic calibration method between LiDAR and cameras \bl{involves the} use of a checkerboard-like target and \bl{the process of} minimizing the re-projection error of the correspondences between two sensor data \cite{ahmad2017extrinsic, zhou2014new,
krishnan2017cross}. Some studies adopted a modified target such as a circular hole \cite{pervsic2017extrinsic, alismail2012automatic}. \bl{However, such} calibration methods, require a target and the performance of the calibration depends on the accuracy of the correspondence estimated from each sensor data. \bl{In particular} when \ac{LiDAR} measurements are non-continuous, the detection of an accurate correspondence between the two modalities is not straightforward. Furthermore, this type of calibration is \bl{not possible} if the two sensors \bl{do not} provide covisibility at \bl{a given moment in time}. \bl{For the purpose of} general calibration, \bl{several} algorithms were introduced to perform calibration using information on the surrounding environment from sensors without using a specific target
\cite{taylor2013automatic, pandey2015automatic, iyer2018calibnet,
zhao2007efficient, tamas2013targetless}. \bl{However}, these approaches rely on an overlapping FOV between two sensors. \bl{In order to overcome} the non-overlapping configuration issue, a local 3D pointcloud is accumulated using the odometry to ensure \bl{that the} overlapping region appears on the camera \cite{scott2015exploiting,
scott2016choosing, napier2013cross}. Another line of study
\cite{taylor2016motion, andreff2001robot, heller2011structure}, \bl{referred to as} hand-eye calibration, estimates the motion of each sensor and performs extrinsic calibration based on the estimated motion relationship.

\begin{figure}[t!]
  \centering
  \subfigure[Pointcloud data from LiDAR]{
	  \includegraphics[width=0.95\columnwidth] {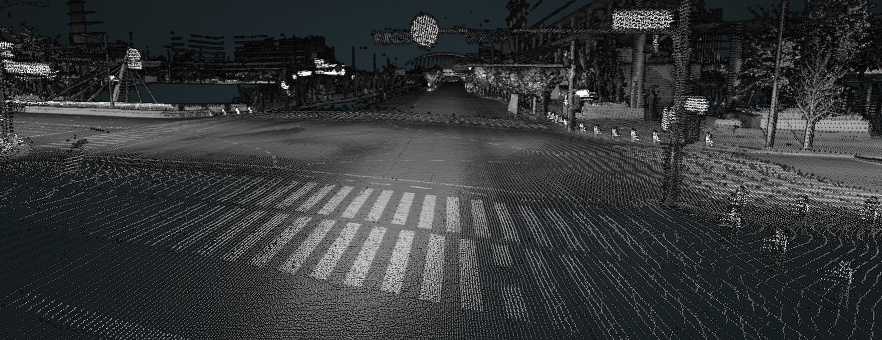}
	  \label{fig:pointcloud}
  }\\
  \subfigure[Projected pointcloud onto a stereo image using calibration result.]{
	  \includegraphics[width=0.95\columnwidth] {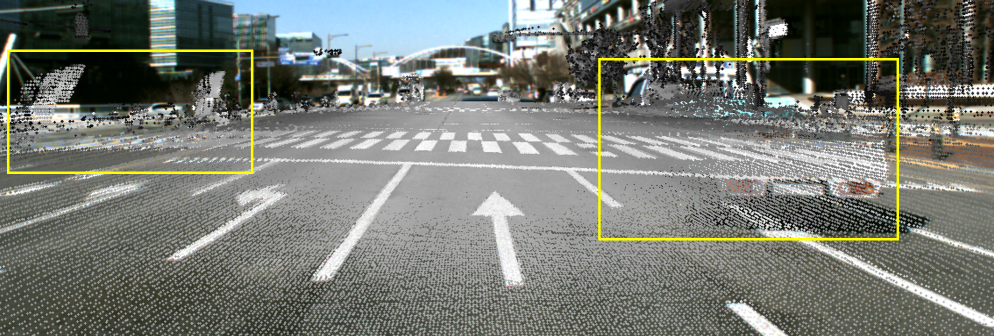}
	  \label{fig:overlap_result}
  }
  \caption{Result of extrinsic calibration between a stereo camera and LiDARs. The projected pointcloud in the image plane coordinate system presents well-matched sensor modalities. The yellow boxes show the inconsistency of the data due to the time difference in the data acquisition.
  }
  \label{fig:main_figure}
  \vspace{-4.00mm}
\end{figure}

\begin{figure}[t!]
  \centering
  \includegraphics[width=0.98\columnwidth] {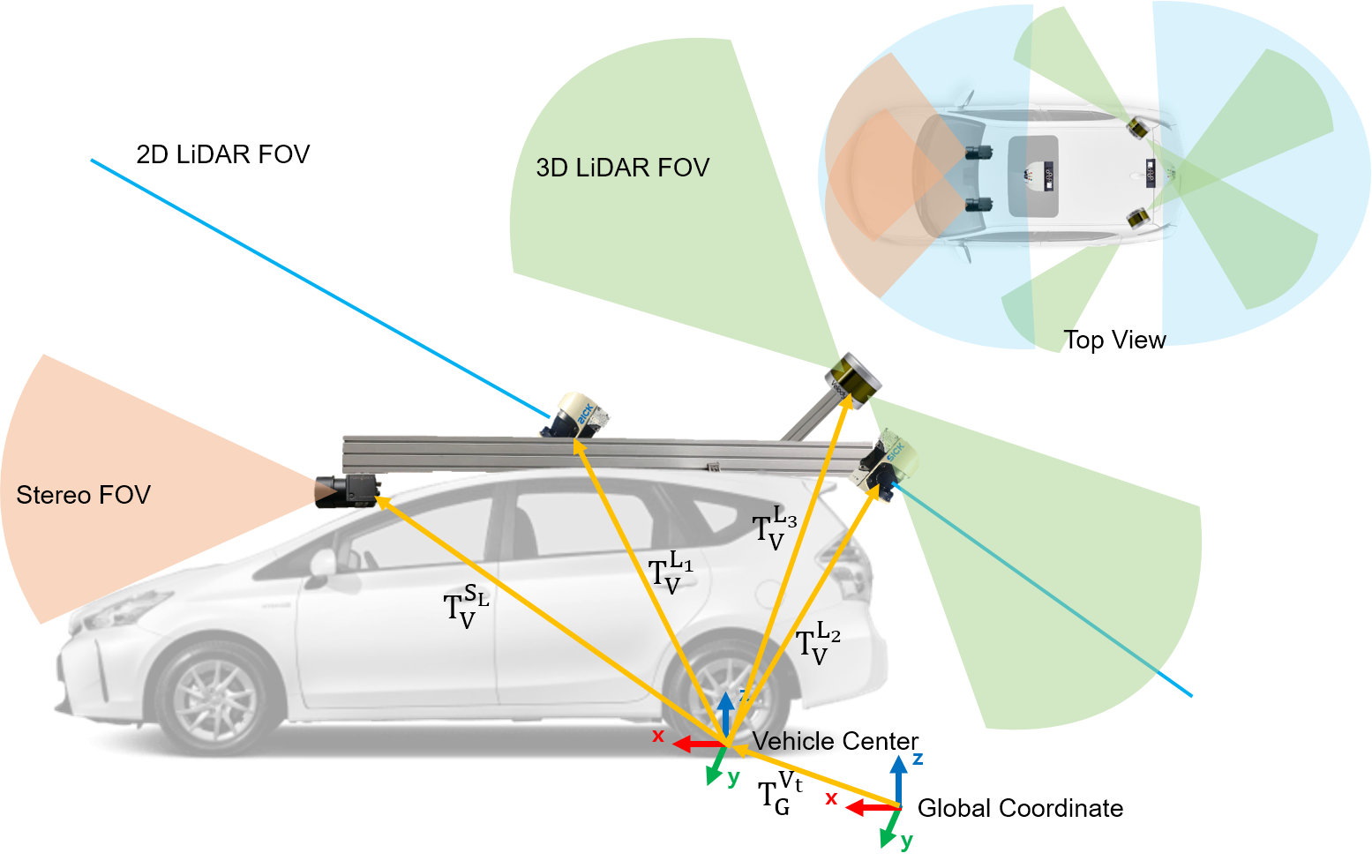}
  \caption{System configuration of the target mobile mapping system. Extrinsic calibration between stereo cameras and LiDAR sensors is a challenging task when there is no guaranteed covisibility.}
  \label{fig:system_configuration}
\end{figure}


The configuration of our sensor system for extrinsic calibration \bl{consist of four LiDARs and a single stereo camera, as shown in} \figref{fig:system_configuration}. Two 3D LiDARs are mounted on the left and right to maximize the range of data acquisition. The two 2D LiDARs are installed facing forward and backward, respectively. \bl{As the} four LiDARs have overlapping regions, calibration was performed using \bl{the geometric information from LiDARs}. However, as the stereo camera is \bl{facing} forward, the images from the stereo camera possess no overlap with any of the LiDAR sensor data.



\section{Related Works}
\label{related_works}

In this paper, we perform an extrinsic calibration between a stereo camera and LiDARs when no overlapping FOV is guaranteed. Literature sources related to our extrinsic calibration method can be divided into two major categories: \bl{target-less and calibration with non-overlapping configurations.}

\subsection{Target-less Extrinsic Calibration}

Target-less extrinsic calibration refers to the method of performing calibration using surrounding data from a general space \bl{without the use of} a checkerboard or a specific shape target. Establishing correspondences between heterogeneous sensor data without a specific target is a highly challenging \bl{task}.

\citeauthor{zhao2007efficient}~\cite{zhao2007efficient} manually selected the
correspondence between LiDAR intensity image and camera image to calculate the
extrinsic parameter. LiDAR intensity image was generated by fusing LiDAR scan
data and odometry estimated using horizontal 2D LiDAR data. \bl{However,} the selection
process for correspondence between heterogeneous sensor data is a laborious task and
\bl{potential source of human errors}. \bl{In order to} mitigate this issue, many
researchers have attempted to calibrate without correspondence \bl{through the use of}
information theory. \citeauthor{taylor2013automatic}~\cite{taylor2013automatic}
proposed to find the optimal alignment between sensors by introducing the \ac{NMI}
metric. \ac{NMI} is a measure of the mutual dependence of \bl{data from} two sensors. This
\bl{study} yields the optimal solution by finding the largest \ac{NMI} value using the
particle swarm optimization method. Similarly, \citeauthor{pandey2015automatic}~\cite{pandey2015automatic} used
the \ac{MI} metric to estimate the extrinsic parameter. \bl{Several local optimal points are encountered if a
single scan is used for optimization.} \bl{In order to} cope
with the problem, \bl{the method of this study} used multiple scan data to find global optimal
points \bl{with} maximum \ac{MI} value. Recently, \bl{with more studies} using deep learning, there have been attempts to perform extrinsic calibration
using deep learning \bl{technique}. \citeauthor{iyer2018calibnet}~\cite{iyer2018calibnet} used the CNN architecture to estimate
calibration parameters using a supervised method.

\subsection{Extrinsic Calibration of Non-overlapping Configuration}

\begin{figure*}[t!]
  \centering
  \includegraphics[trim={2.5cm 5.8cm 2.2cm 5.0cm},clip,width=0.98\textwidth]{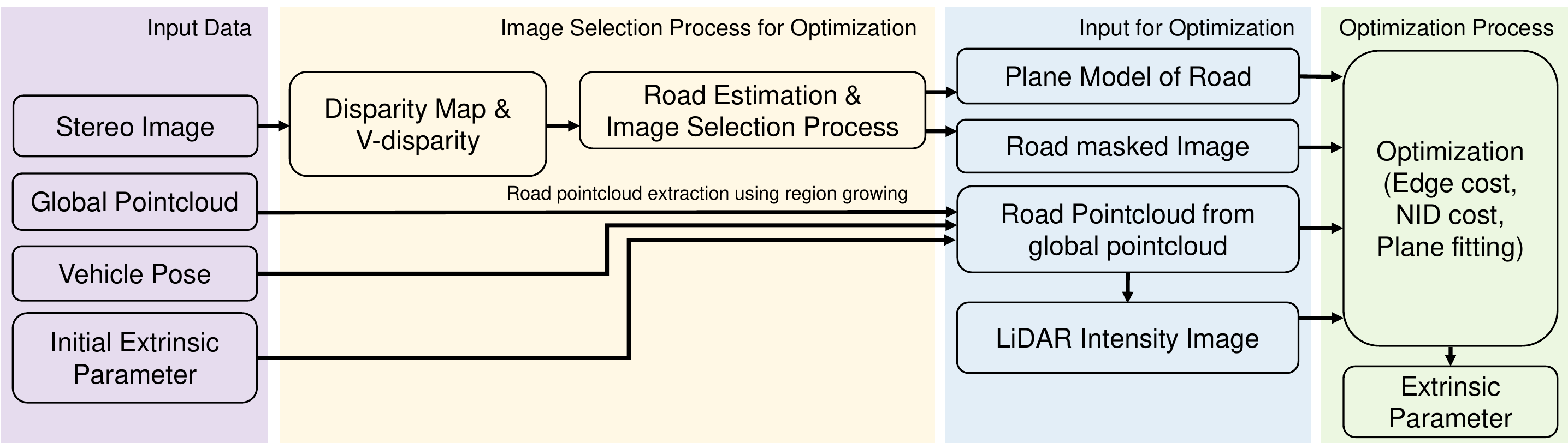}
  \caption{Overall process of automatic calibration. The colored boxes refers to the different processes. Details of the image selection process are described in Fig~\ref{fig:image_selection_diagram}.}
  \label{fig:whole_diagram}
\end{figure*}

The aforementioned target-less extrinsic calibration still requires an
overlap between sensors. \bl{Several} studies have focused on solving extrinsic
calibration for configurations without any overlap between sensors. For
example, \citeauthor{napier2013cross}~\cite{napier2013cross} performed extrinsic calibration between
push-broom 2D LiDAR and cameras. The push-broom LiDAR \bl{obtained} distance measurement
from the \bl{left-hand and the right-hand sides} of the vehicle, and the cameras obtained 
images from the front. \bl{This} means that there was no overlapping region between the two
sensor data. Therefore, extrinsic calibration was performed using
edge images from the camera image and LiDAR intensity image generated by
projecting the 3D pointcloud \bl{that was} calculated by applying the vehicle odometry
measurement.

\citeauthor{scott2015exploiting}~\cite{scott2015exploiting} leveraged the \acf{NID}
metric to estimate optimal extrinsic parameter between 2D LiDAR and multiple
cameras. In the optimization process, they utilized \bl{factory-calibrated}
transformation as constraint. \citeauthor{scott2016choosing}~\cite{scott2016choosing} added a data selection
module by comparing the \ac{NID} value around the optimal parameter to
reduce the local optimal points.

Our proposed method utilizes the road marking information for the extrinsic calibration
of non-overlapping configurations. The road are observed together \bl{through} both sensors (camera and LiDAR)
regardless of the acquisition time of the data. \bl{In order to} minimize the local optimal point, multiple images were used and the images were selected using the costs defined with the vanishing point. In the optimization process, we defined multiple costs using image and LiDAR data which \bl{include} \ac{NID} cost, edge alignment cost and plane fitting cost to make it more convergent. This paper presents the following:

\begin{itemize}
	\item An automatic extrinsic calibration framework for non-overlapping configuration
	\item Utilization of road regions estimated using stereo images for a more stable optimization process
	\item An automatic image selection process using estimated vanishing point and road markings
	\item A multiple cost function for robust optimization even with rough initial values
\end{itemize}

\section{Notation and Local Map Generation}
\label{notation_localmap}

We first generated a 3D pointcloud using odometry and LiDAR data and performed
calibration by comparing the stereo images and LiDAR intensity image over the
generated local pointcloud map.  For local map generation, the odometry of
the vehicle was calculated \bl{with high accuracy using} the wheel encoder, the 3-axis \ac{FOG}, and the \ac{IMU}. Before describing the details of the extrinsic calibration, \bl{this paper} first
introduces notations and assumptions for generating the local map.

\subsection{Notation}

Matrix $T_B^A \in SE(3)$ represents the rigid body transformation that registers the data defined in $A$ coordinate system to $B$ coordinate system. Matrix $T_B^A$ is also represented by a tuple $t_B^A \in \mathbb{R}^6$, where $t_B^A = \{t_x, t_y, t_z, r_x, r_y, r_z\}$. Here, $t_x, t_y$ and $t_z$ represent the relative translation along each axis in meters, and $r_x, r_y$ and $r_z$ represent the relative rotation that is roll, pitch, and yaw in radians. ${}^CP_D$ represents the pointcloud set from sensor $D$ in coordinate system $C$. For instance, ${}^{V_t}P_{L_{i,t}}$ represents the pointcloud from the $i_{th}$ LiDAR at time $t$ in the coordinate system $V$. Subscripts $L, S_L, V$, and $G$ denote LiDAR, the left camera of stereo, the vehicle center and the global coordinate system.

\subsection{Local Map Generation}

\bl{Due to the fact that} no overlap \bl{is attained} between the camera and the LiDAR at \bl{any given instance of time}, we accumulated LiDAR data to generate a local pointcloud map. We assume that
accurate relative transformation is calculable using a 3-axis \ac{FOG}, an \ac{IMU} and
wheel encoder. We also began with pre-computed extrinsic calibration between
LiDARs and LiDAR-to-vehicle using our \bl{previously proposed} algorithm \cite{jjeong2018icra}.

Given an accurate relative sequential pose and extrinsic calibration between the four
\ac{LiDAR}s, we accumulated the LiDAR data and constructed a local pointcloud map
written in the global coordinate system. The local pointcloud ${}^{L_{i}} P_{L_{i, t}}$
\bl{that was} written in each sensor coordinate system at time $t$ was represented in the
vehicle coordinate system using the computed vehicle-to-LiDAR coordinate system transform
$t_{V}^{L_i}$ as \bl{described} in \eqref{eq:lidar2vehicle}. \bl{This was followed by the sequential transformation of pointcloud
${}^{V_t}P_{L_{i,t}}$ from each LiDAR sensor in the vehicle coordinate system} into the global coordinate frame via
\eqref{eq:lidar2global}. Lastly, the global pointcloud ${}^GP_L$ was obtained by
adding of the all pointcloud of each LiDAR \bl{via} \eqref{eq:lidar_sum}.

Our method used the global pointcloud ${}^GP_L$, the stereo image $I_{S_L}$, and
the odometry $t_{G}^{V_t}$ for stereo extrinsic calibration. \bl{Particular attention} was required
as the \bl{accuracy of the local pointcloud map} depends on the accuracy of the relative
vehicle pose estimation. In this \bl{study}, we accumulated the local map for
\bl{a range of approximately \unit{80}{m}} to minimize the potential accumulation error.
\begin{eqnarray}
  {}^{V_t}P_{L_{i,t}} &=& t_{V}^{L_i} \oplus {}^{L_{i}} P_{L_{i, t}} \label{eq:lidar2vehicle}\\
  {}^GP_{L_{i,t}} &=& t_{G}^{V_t} \oplus {}^{V_t}P_{L_{i,t}} \label{eq:lidar2global}\\
  {}^GP_L &=& \sum\limits_{i=1}^{n} \sum\limits_{t=0}^{T}{}^GP_{L_{i,t}} \label{eq:lidar_sum}
\end{eqnarray}
The local pointcloud map was obtained at time $t$ when the \ac{LiDAR} measurement
was received. As the camera images were captured at different \bl{frequencies}, we
needed to consider the local pointcloud at time $t+k$ when the image is received.
The pointcloud in the global coordinate system can be transformed into the vehicle
coordinate system using the odometry information $t_{V_{t+k}}^{G}$ at time $t+k$
time at which the stereo image was acquired \eqref{eq:global2vehicle}. The
pointcloud ${}^{V_{t+k}}P_L$ in the vehicle coordinate system was \bl{subsequently} registered into the
stereo camera coordinate system using the extrinsic parameter $t_{S_L}^{V}$ of
the stereo camera, \bl{which is what our method is aiming to calculate.} \eqref{eq:vehicle2stereo}. The
\ac{LiDAR} intensity image was generated by projecting the pointcloud onto stereo
image plane (\figref{fig:main_figure}) using the pre-calibrated intrinsic parameter $K_{S_L}$
\eqref{eq:intrinsic_projection}.
\begin{eqnarray}
  {}^{V_{t+k}}P_L &=& t_{V_{t+k}}^{G} \oplus {}^GP_L \label{eq:global2vehicle} \\
  {}^{S_{L, t+k}}P_L &=& t_{S_L}^{V} \oplus {}^{V_{t+k}}P_L \label{eq:vehicle2stereo}\\
  {}^{S_{L, t+k}}I_L &=& K_{S_L} {}^{S_{L, t+k}}P_L \label{eq:intrinsic_projection}
\end{eqnarray}

\section{Target-less Multi-modal Extrinsic Calibration with No Overlap}
\label{proposed_method}

\bl{With the} constructed local map, the proposed method peforms two
steps: (\textit{i}) informative image selection and (\textit{ii}) multiple cost
optimization. In the image selection process, the road regions are extracted using
a stereo disparity map, and images that include abundant road markings were
selected \bl{for the optimization process by extracting the lines facing the vanishing point.} \bl{The local optima issue can be alleviated by} using multiple images to estimate extrinsic calibration.
\figref{fig:whole_diagram} shows the overall
calibration process. \bl{The following sections} describe each module in detail.


\begin{figure}[t!]
  \centering
  \includegraphics[trim={8cm 0.5cm 3cm 0.5cm},clip,width=0.98\columnwidth]{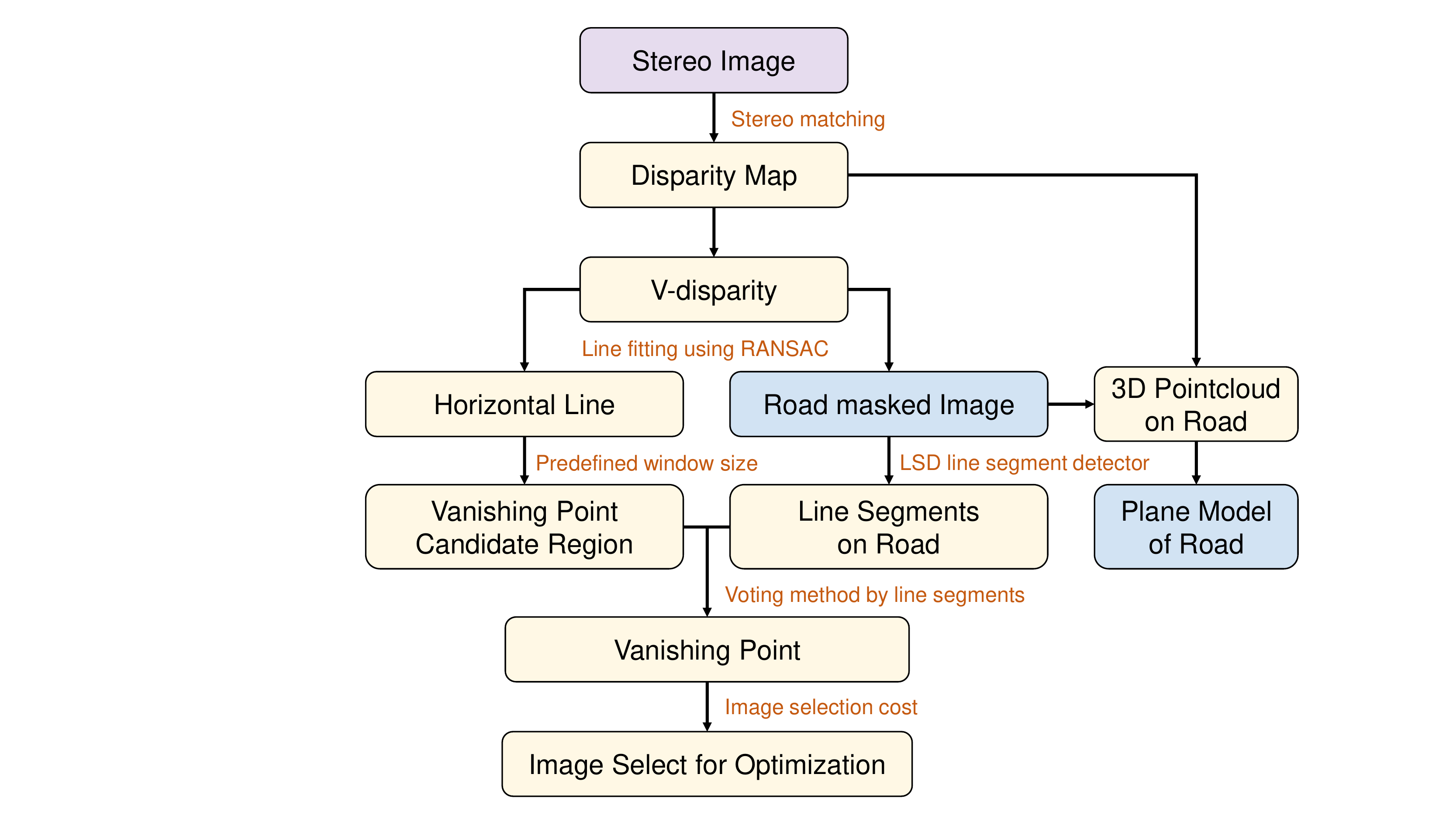}
  \caption{Image selection process using stereo images.}
  \label{fig:image_selection_diagram}
\end{figure}

\subsection{Road Detection and Informative Image Selection}

\bl{Images are marked as} informative if \bl{abundant} road markings are contained. In
this section, we detect the road region within an image and propose the image
utility measure for extrinsic calibration. The overall scheme is \bl{provided} in
\figref{fig:image_selection_diagram}.

\subsubsection{Road Region Detection using Stereo Images}
\label{sec:road_region}

\begin{figure}[!t]
  \centering
  \subfigure[Left image of the stereo camera and v-disparity]{
	  \includegraphics[width=0.98\columnwidth] {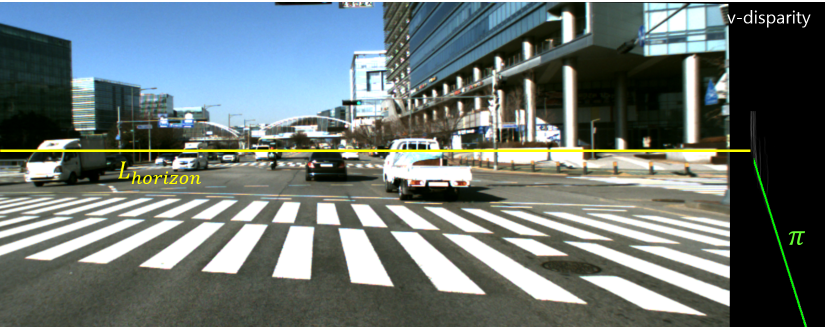}
	  \label{fig:image_with_v_disparity}
  }\\
  \subfigure[Disparity map $D(p_{uv})$ of stereo image]{
	  \includegraphics[width=0.98\columnwidth] {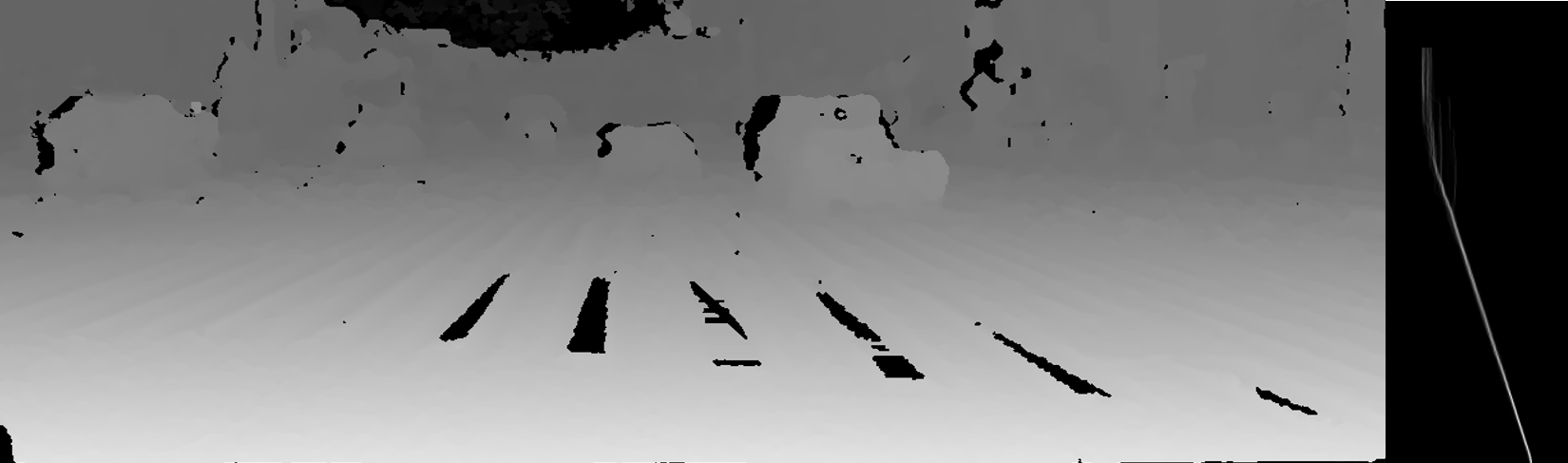}
	  \label{fig:stereo_disparity_map}
  }\\
  \subfigure[Road mask extracted by plane fitting in v-disparity]{
	  \includegraphics[width=0.98\columnwidth] {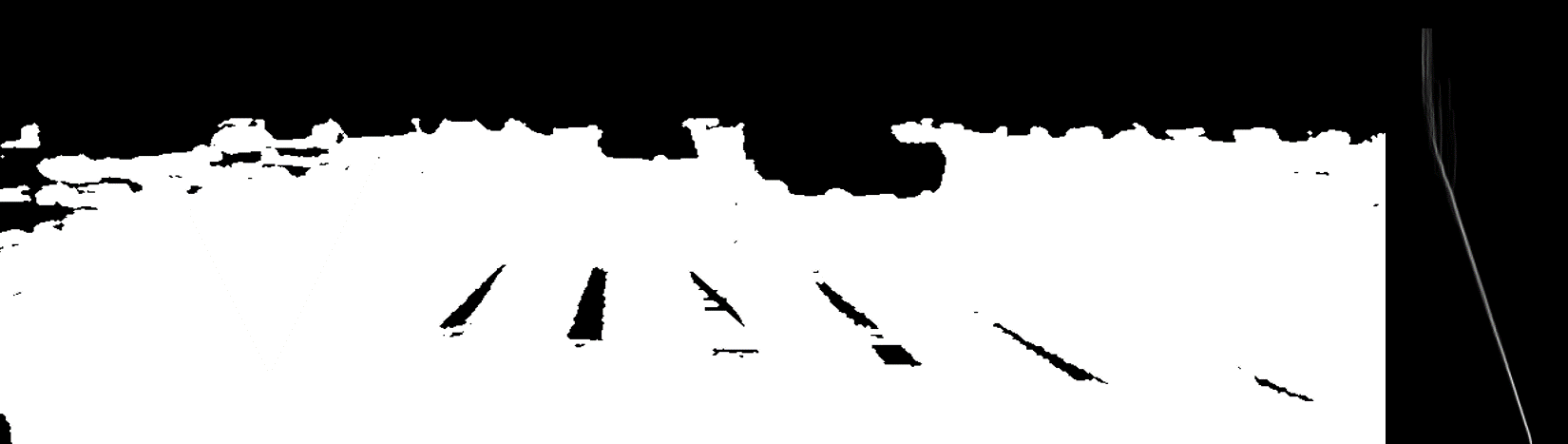}
	  \label{fig:road_mask}
  }\\
  \caption{Road extraction process using a stereo image. The disparity map is estimated via stereo matching and the v-disparity is calculated through projection onto the plane of (v, disparity). The $\pi$ in the v-disparity refers to the road surface.}
  \label{fig:road_extraction_process}
  \vspace{-8.00mm}
\end{figure}

In the proposed method, we compute the v-disparity image \cite{labayrade2002real} from a
disparity map $D(p_{uv})$ for a pixel $p_{uv}$ to detect the road region. The
histogram of the row line on the disparity map is depicted on the row of the v-disparity
image, and sample v-disparity images \bl{are shown} on the right-hand side of
\figref{fig:road_extraction_process}.

\bl{Following this process,} the road region \bl{is detected} by fitting a line $\pi$ on the v-disparity image
using \ac{RANSAC}. \bl{Due to the fact that} each pixel value of the v-disparity image \bl{denotes} the
histogram of the disparity map, fitting a line captures the continuously
changing disparity \bl{along the $v$ direction (i.e., ground) of the image.} The road
region \bl{is informed} from pixels corresponding to the fitted line $\pi$. The sample of the detected road region can be seen in \figref{fig:road_mask}.



\subsubsection{Vanishing Point Estimation}

\begin{figure}[!b]
  \centering
  \includegraphics[width=0.49\textwidth]{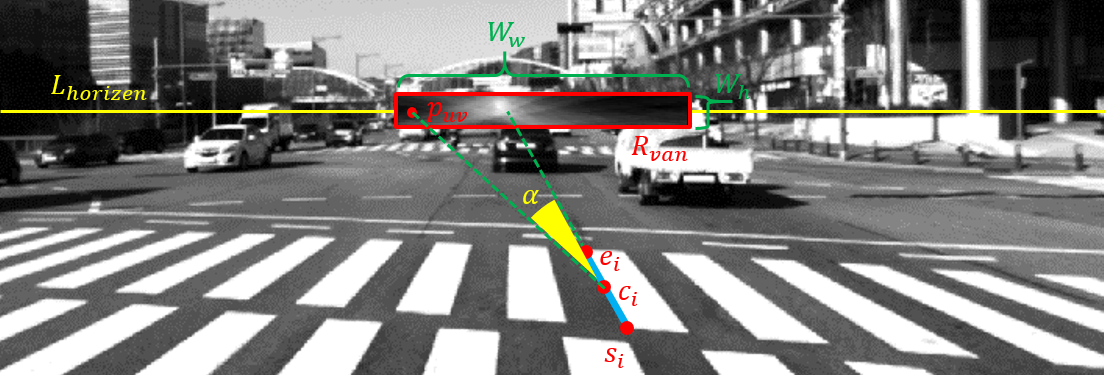}
  \caption{Voting process for vanishing point estimation. The yellow line refers to the horizon calculated using v-disparity and the red box refers to the candidate vanishing point region. Each line extracted from the road image votes on every pixel in the candidate area.}
  \label{fig:vanishing_region}
\end{figure}

Next, \bl{the vanishing point is estimated} from the detected road region. The vanishing
point allows \bl{for the discerning of} road markings from other features such as the shadows of
street lamps and trees. Given the line $\pi$ in the v-disparity image
(i.e., the road plane), the $v$ value at which $\pi$ meets the $v$-
axis \bl{is identified}. The horizontal line at this $v$ value \bl{is referred to} as the $L_{horizon}$, as depicted in
\figref{fig:image_with_v_disparity}.


The $L_{horizon}$ is near to the vanishing point but may not be exact. \bl{A voting process is additionally applied} to estimate the vanishing point near
the $L_{horizon}$. After defining a voting region $R_{van}$ of size $W_w$ by $W_h$
around the central point of $L_{horizon}$, the lines $\widehat{L_{S_L}} = \{l_1,
l_2, \cdots , l_n \}$\bl{, which are voters participating in the vote, were} detected on the road region using a \ac{LSD}
\cite{von2010lsd}. The symbol
$\widehat{\cdot}$ \bl{is used} to indicate that the corresponding data was extracted from the road
region. Each line $l_i = \{s_i, e_i, c_i \}$, where $s_i, e_i$ and $c_i$ are the
start, end and center point of each line respectively, votes depending on the weight
over each pixel $p_{uv}$ in the $R_{van}$ \eqref{eq:voting}.
\begin{eqnarray}
  \label{eq:voting}
  \text{Vote}(p_{uv}) =
  \begin{dcases}
  \min \Bigl( \frac{1}{\alpha},10 \Bigr)  &  \text{if } \alpha \leq \rho_{1}\\
  0 & otherwise
  \end{dcases}
\end{eqnarray}
\begin{eqnarray}
  p_{van} &=& \argmax_{p_{uv}} R_{van}(p_{uv}) \label{eq:vanishing_point_calculate}\\
  U_{van} &=& \max R_{van}(p_{uv}) \label{eq:vanishing_poit_certenty}
\end{eqnarray}

\begin{figure}
  \centering
  \includegraphics[width=0.98\columnwidth]{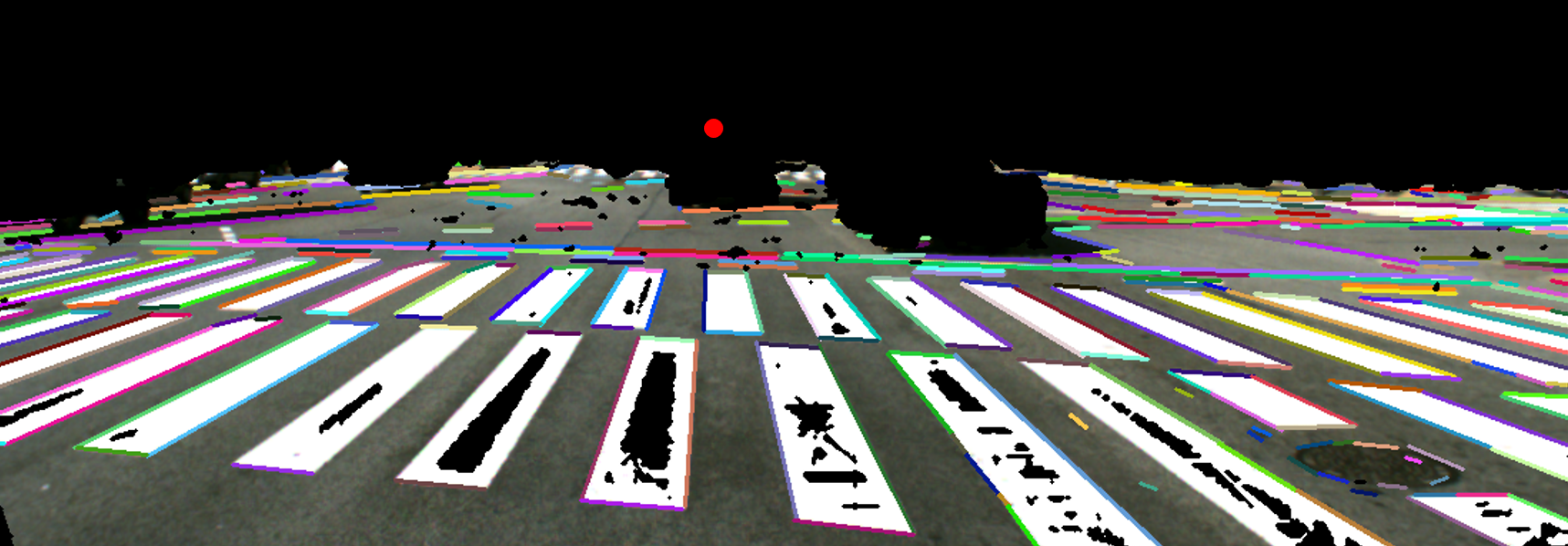}
  \caption{Result of the vanishing point extraction process. The red point represents the estimated vanishing point. The lines extracted from the road are voters that participate in the vote.}
  \label{fig:vanishing_point_result}
\end{figure}

Here, $\alpha = \angle (\overline{p_{uv},c_{i}},\overline{s_{i},e_{i}})$ refers to
an angle between voter $l_i$ and the line connecting $c_i$ and $p_{uv}$ in degrees, as
illustrated in \figref{fig:vanishing_region}. \bl{The voting weight of each line} is the
inverse of the angle to emphasize lines with smaller angles (with a maximum
weight of 10). In this paper, threshold $\rho_{1}$ was set to $3^\circ$. The
vanishing point $p_{van}$ was determined as the largest value in the voting
results \eqref{eq:vanishing_point_calculate}, and $U_{van}$ denotes the
corresponding confidence level. \figref{fig:vanishing_point_result} shows the estimated vanishing point and line segments on the road. 


\subsubsection{Image Selection}

The utility of each image $U_I$ for image selection
is calculated using $p_{van}$, $\widehat{L_{S_L}}$ and $U_{van}$
\eqref{eq:image_selection_utility}. A higher $U_I$ represents a greater number of
road markings, \bl{which in turn indicates} better data for calibration. In order to suppress the effect of
the local optimal point in the optimization process, multiple images and the
corresponding pointcloud were selected. \bl{Based on the utility, five
images with the highest utility are selected.}

\begin{equation}
  \label{eq:image_selection_utility}
  U_I =  \Bigl( \sum\limits_{i=1}^N \min \Bigl( \frac{1}{\angle (\overline{p_{van},c_i}, \overline{s_i,e_i})},1 \Bigr)  \Bigr) \cdot U_{van}
\end{equation}

\subsubsection{Plane Estimation of the Road}

Once the informative images are selected, the plane $M=\{n_x, n_y, n_z, d\}$, is
estimated from the disparity map $D(p_{uv})$. \bl{Conversions are made} from $D(p_{uv})$ in
camera coordinate system to the 3D pointcloud using the camera model: the focal length $f$, and
base line $B$. $M$ is subsequently calculated using a \ac{RANSAC} plane fitting with the
pointcloud. The plane is used in optimization process to calculate the plane
fitting cost.



\begin{figure}[!t]
  \centering
  \subfigure[The LiDAR intensity image $\widehat{I_L}$ (left) and edge image $\widehat{E_L}$ extracted from LiDAR intensity image (right)]{
	  \includegraphics[width=0.98\columnwidth] {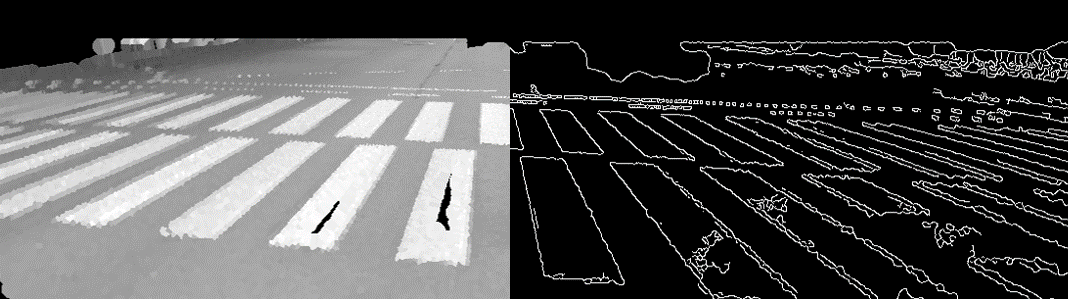}
	  \label{fig:lidar_intensity_image}
  }\\
  \subfigure[Edge image $\widehat{E_{S_L}}$ extracted from stereo image (left) and distance transform image $\widehat{G_{S_L}}$ using inverse of edge image from stereo (right)]{
	  \includegraphics[width=0.98\columnwidth] {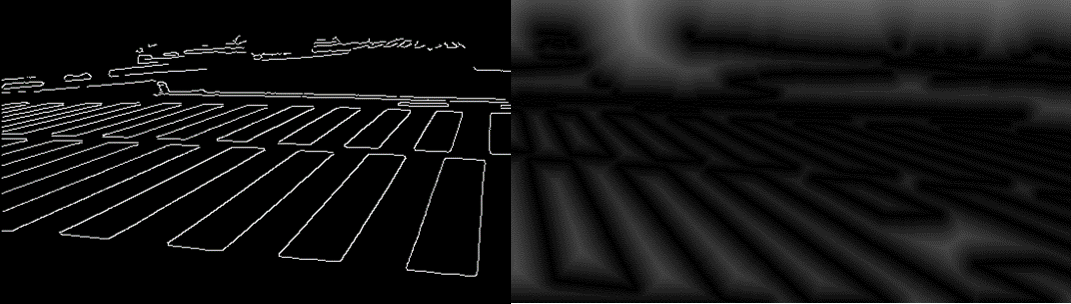}
	  \label{fig:edge_img_from_intensity_image}
  }\\
  \caption{The process of computing the edge alignment cost. The edge is extracted using the canny edge algorithm with the LiDAR intensity image and the stereo image. The distance transformed image is calculated using the inversely transformed edge image of the stereo image.}
  \label{fig:estimate_distance_transform}
\end{figure}

\subsection{Multi-cost Optimization}


Through the aforementioned image selection process, a \ac{LiDAR} intensity image,
a road masked image, and a plane model were obtained. For the extrinsic calibration,
we also extracted points belonging to roads. Specifically, the road pointcloud
${}^{S_L}\widehat{P_L}$ was calculated from the global pointcloud using the region
growing segmentation method \cite{rabbani2006segmentation}. For extrinsic
calibration, this paper proposes using the following three costs.

\begin{figure*}[!t]
  \centering
  \subfigure{
	  \includegraphics[height = 0.87\columnwidth] {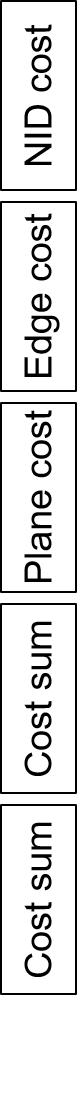}
  }
  \subfigure{
	  \includegraphics[trim={0cm 2.5cm 1cm 0.5cm},clip,width=0.92\textwidth]{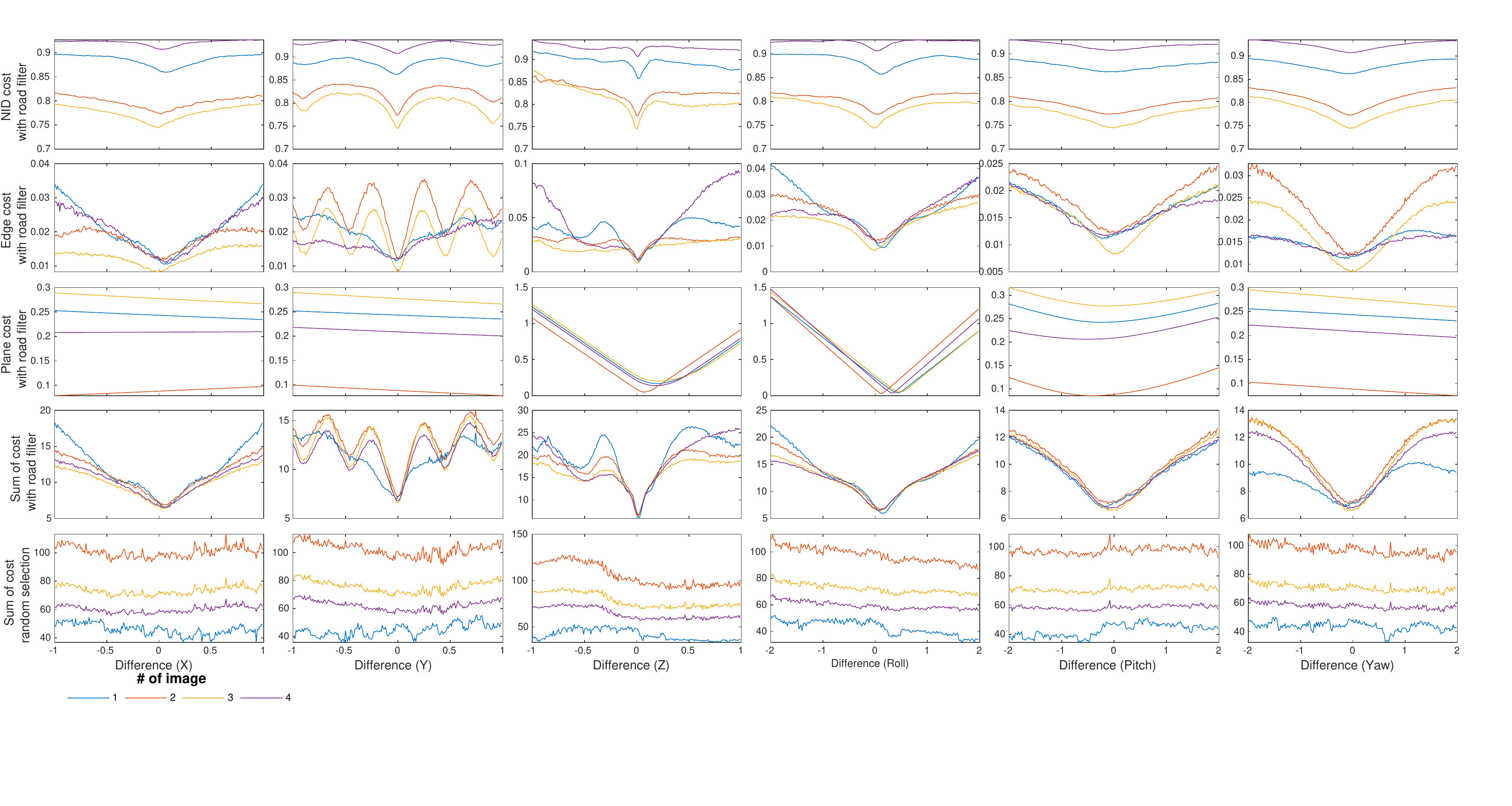}
  }
  \caption{Variation of each cost according to parameter changes based on the optimal extrinsic parameter. The first row to the fourth row shows the change in cost according to the number of images for the selected image and road filtering. The fifth row shows the cost change when the \bl{images are selected randomly}. The units for translation and rotation are meter and degree respectively. }
  \label{fig:cost}
\end{figure*}

\subsubsection{Edge Alignment Cost}

The first cost, edge alignment cost, is used to evaluate the discrepancy between RGB and LiDAR
intensity images. A camera edge image ($\widehat{E_{S_L}}$) is
extracted from a road masked image ($\widehat{I_{S_L}}$), and a \ac{LiDAR} edge image
($\widehat{E_{L}}$) is computed from a \ac{LiDAR} intensity image
($\widehat{I_{L}}$) by applying the canny edge algorithm
\cite{canny1986computational}.

In order to compare the two edge images by defining a differentiable cost function, RGB-induced edge image is converted into a distance transform image $\widehat{G_{S_L}}$. The edge alignment cost was computed by calculating the sum of the multiplied pixel value of $\widehat{E_{L}}$ and $\widehat{G_{S_L}}$
\eqref{eq:edge_cost}. Defining the cost using the distance image aids the algorithm to
converge with rough initial parameters. The edge cost reaches a minimum value when
the two data precisely match one another. \figref{fig:estimate_distance_transform} shows
the process of calculating the edge alignment cost.
\begin{eqnarray}
  \label{eq:edge_cost}
  f_{edge} (\widehat{E_{L}}, \widehat{G_{S_L}}) = \sum\limits_{p_{uv}}  \widehat{E_{L}}(p_{uv}) \widehat{G_{S_L}}(p_{uv})
\end{eqnarray}

\subsubsection{NID Cost}

The second cost is induced from the \ac{NID} metric, which measures the correlation
between two random variables, $X$ and $Y$. In this paper, the \ac{NID} cost between
$\widehat{L_{S_L}}$ and $\widehat{I_L}$ was used to identify the best alignment. The
\ac{NID} cost was defined as:
\begin{eqnarray}
  f_{NID} (\widehat{I_{S_L}}, \widehat{I_L}) &=& 2 - \frac{H(\widehat{I_{S_L}}) + H(\widehat{I_L})}{H(\widehat{I_{S_L}},\widehat{I_L})} \label{eq:NID_cost}\\
  H(X) &=& - \sum\limits_{x \in X} P(x) \log P(x) \nonumber\\
  H(X,Y) &=& - \sum\limits_{\substack{x \in X\\y \in Y}} P(x,y) \log(P(x,y)). \nonumber
\end{eqnarray}
Here, $H(X)$ and $H(X,Y)$ are single entropy and cross entropy. The computed
\ac{NID} metric has the range of $0 \leq f_{NID} (\widehat{I_{S_L}}, \widehat{I_L}) \leq
1$. Lower \ac{NID} costs indicate a greater degree of similarity between the distributions of
two data. In our case, the intensity value of \ac{LiDAR} and the image pixel value in
grayscale achieves the best alignment between two images with the lowest NID
cost.

\subsubsection{Plane Fitting Cost}

The final cost is computed using the plane model $M$ obtained from the stereo
image and the road pointcloud ${}^{S_L}\widehat{P_L}$ acquired from the global
3D map. By including this geometric cost, convergence of the $z$ and the rotation
parameter is improved.
\begin{equation}
  \label{eq:plane_fitting_cost}
  \begin{multlined}
  f_{plane}({}^{S_L}P_L, M) = \\ \sum\limits_{i=1}^{N}\Bigl( n_x  ({}^{S_L}P_L)^i_x + n_y  ({}^{S_L}P_L)^i_y + n_z ({}^{S_L}P_L)^i_z + d\Bigr)
  \end{multlined}
\end{equation}

In this cost, $({}^{S_L}P_L)^i_x$, $({}^{S_L}P_L)^i_y$ and $({}^{S_L}P_L)^i_z$
are the $x$, $y$, and $z$ values of $i^{\text{th}}$ point in the stereo camera coordinate system.
The cost is summed over the number of the points ($N$) in the target pointcloud
${}^{S_L}P_L$. Lower plane fitting costs are yielded when the pointcloud and
road model closely match one another.

\subsubsection{Optimization}

Optimization is performed using the weighted sum of the three costs:
$f_{edge}$, $f_{NID}$ and $f_{plane}$. In this paper, $k_1$, $k_2$, and $k_3$ are
set as 2.0, 500.0, 0.1 respectively, considering the scale of each cost.

\begin{equation}
\label{eq:total_cost}
f_{sum} = k_1 f_{edge} + k_2 f_{NID} + k_3 f_{plane}
\end{equation}

\begin{equation}
\label{eq:final_optimization}
t^V_{S_L} = \argmin_{t^V_{S_L}} f_{sum}
\end{equation}

The extrinsic parameter of the stereo camera $t^V_{S_L}$ was estimated by
identifying the minimum point of sum of the costs by utilizing the downhill simplex method
\eqref{eq:final_optimization}.

\section{Experimental Results}
\label{result}

This section presents experimental results to evaluate the defined cost in both
a quantitative and qualitative manner. We also verified the repeatability and
sensitivity of the extrinsic parameter between the stereo camera and \ac{LiDAR}
during the proposed extrinsic calibration.

\subsection{Evaluation of Cost Function}

We started by depicting the cost function with respect to the perturbation from
the optimal values. The variation of each \bl{cost} and total cost are plotted in
\figref{fig:cost} by increasing the number of selected images. From the first
row to the fourth row, up to four informative images were incrementally included
in the information order\bl{,} from the most informative to the fourth \bl{most} informative
image. For these cases, the cost was calculated using only the information on
the road region.

As \bl{shown in the results}, the proposed cost reshapes the function in all parameter spaces.
The NID cost has a smooth curve shape compared to the edge cost, \bl{whereas} the
segment for converging to the global optimal point is relatively narrow. \bl{On the other hand,} the
edge cost \bl{graph} is somewhat noisier but has a broader segment to
converge to the global optimal point. This feature of the edge cost helps to
converge to the global optimal point even \bl{with a large} initial error of calibration. The shape of the edge cost \bl{graph} in the y-direction (lateral direction) can be
explained considering that the selected informative images tended to contain more
road markings with repeated patterns (e.g., cross walks). However, note that the
NID cost alleviated this repeated local minimum in the y-axis direction in the
summed cost function.

The plane fitting cost affects z, roll, and pitch by exploiting the \bl{geometric road}
information\bl{, and} mainly enhances the convergence in z and roll. The plane fitting
cost depends on the accuracy of the disparity map. The computed disparity map
may include errors due to pixel discontinuity, thus \bl{resulting in a less sharp shape at the optimal point for} the optimal point. To incorporate the global tendency
while hindering the local details of the plane cost, we set the weight of the
plane fitting cost to be somewhat smaller. Overall, the sum of the cost shows a
distribution of shapes that can converge toward the global optimal point in the
\bl{overall} section.



The graph in the last row reveals the effect of the image selection. The plot
shows the cost variation without an image selection process, by selecting a image randomly.
\bl{When an image is arbitrarily selected, the cost produces significant amount of noise} for all the parameters and fails to provide a clear global optimal point regardless
of the number of images. This proves that an informative image selection
\bl{process} significantly influences extrinsic parameter estimation.

\begin{table}[t!]
\begin{tabular}{c|cccccc}
 & \begin{tabular}[c]{@{}c@{}}X\\ {[}m{]}\end{tabular} & \begin{tabular}[c]{@{}c@{}}Y\\ {[}m{]}\end{tabular} & \begin{tabular}[c]{@{}c@{}}Z\\ {[}m{]}\end{tabular} & \begin{tabular}[c]{@{}c@{}}Roll\\ {[}degree{]}\end{tabular} & \begin{tabular}[c]{@{}c@{}}Pitch\\ {[}degree{]}\end{tabular} & \begin{tabular}[c]{@{}c@{}}Yaw\\ {[}degree{]}\end{tabular} \\ \hline
Left & 1.669 & 0.278 & 1.612 & -91.124 & -0.632 & -90.236 \\
Right & 1.710 & -0.201 & 1.595 & -90.985 & -0.603 & -90.146 \\
Diff & 0.041 & 0.479 & 0.017 & 0.139 & 0.029 & 0.09 \\
GT & 0 & 0.475 & 0 & 0 & 0 & 0 \\
Error & 0.041 & 0.004 & 0.017 & 0.139 & 0.029 & 0.09
\end{tabular}
\caption{Quantitative evaluation using a stereo camera. There \bl{should} be a translation of the baseline in exactly the y-axis direction between the rectified images.}
\label{tab:result_table}
\end{table}

\subsection{Repeatability of Parameter Optimization}

\begin{figure}[t!]
  \centering
  \includegraphics[trim={0.5cm 0.5cm 1.0cm 0.5cm},clip,width=0.98\columnwidth]{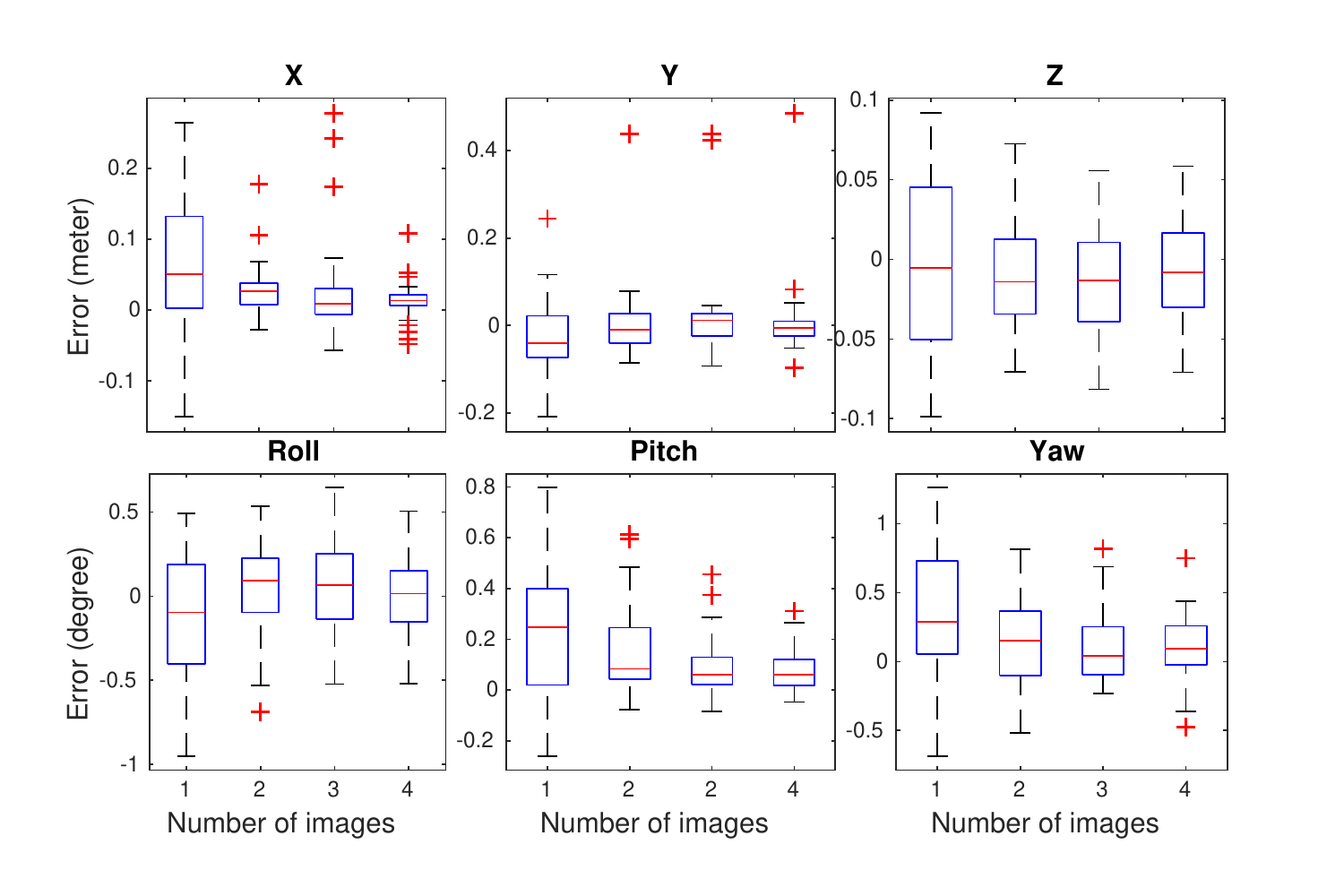}
  \caption{Repeatability of the parameter estimation with respect to the number of images used. The calibration was performed 40 times using random initial points while measuring error against the calibration result of using all five informative images.}
  \label{fig:repeatability_teset}
  \vspace{-4.00mm}
\end{figure}

Repeatability is the property of outputting the same result for different inputs
through repetition of the algorithm. In order to test repeatability, the
algorithm was executed 40 times for each number of images using random initial
values \bl{within} the range of \unit{0.3}{m} and $3^\circ$ for translation and rotation,
respectively (\figref{fig:repeatability_teset}). The top and bottom of the box plot \bl{refer to the} $25^{\text{th}}$ and $75^{\text{th}}$
percentiles respectively, and the middle red line \bl{is} the median of the error.
The higher the number of images, the better the \bl{overall} repeatability as a whole, and
the repeatability is significantly improved when two or more images are used.

\begin{figure}[b!]
  \centering
  \subfigure{
	  \includegraphics[width=0.98\columnwidth] {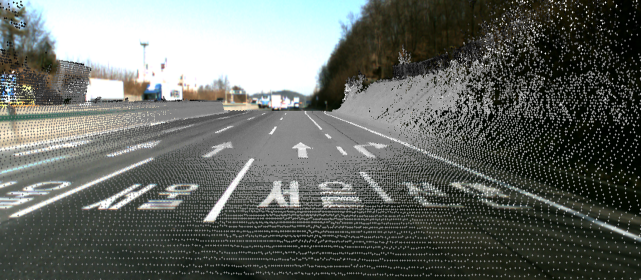}
  }\\
  \subfigure{
	  \includegraphics[width=0.98\columnwidth] {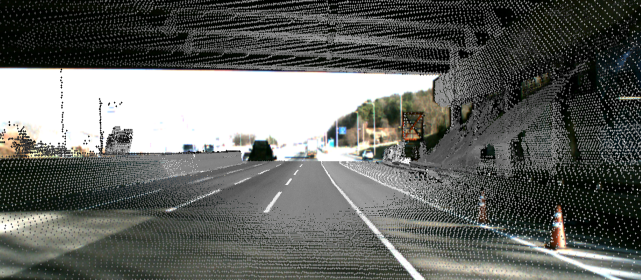}
  }\\
  \caption{Qualitative evaluation of the calibration. We projected the intensity value of the pointcloud onto the stereo image in a highway environment.}
  \label{fig:quality_eval}
\end{figure}

\subsection{Comparison with Available Ground Truth}

It is challenging to calculate the ground truth of the actual calibration
parameters for extrinsic calibration, especially when no overlap is
guaranteed as \bl{in the case of} our system. Therefore, in this paper, we used the baseline of the
stereo camera as the ground truth. If the stereo camera is correctly calibrated,
there \bl{should} be a translation of the baseline in exactly the y-axis direction
between the rectified images, and the rest of the translation and rotation
should be zero.

The extrinsic parameters of \bl{each of the two} rectified images were calculated using the
proposed algorithm, and the difference between the two parameters
was compared. \tabref{tab:result_table} shows the comparison with the ground truth.
On average, there was an error of \bl{approximately} \unit{0.02}{m} for translation and \bl{approximately}
$0.086^\circ$ for rotation.  \figref{fig:quality_eval} shows the image when the
global pointcloud is projected onto the left stereo camera \bl{image} using the finally
computed extrinsic parameter. \bl{We can see} that the pointcloud of the road and
the structure is projected correctly onto the corresponding pixels.

\section{Conclusion and Future Works}
\label{conclusion}

This paper proposed an automatic extrinsic calibration \bl{method} for estimating the rigid
body transformation between a stereo camera and \ac{LiDAR} with no overlapping
\ac{FOV}. By exploiting the static features \bl{in the urban environment such as road markings}, it was confirmed that extrinsic parameters could be calculated
with \bl{a degree of} accuracy of \bl{approximately} \unit{0.02}{m} and $0.086^\circ$ without manual
operation by a human \bl{operator}.

In this paper, we used the assumption that the odometry for generating poitncloud is
locally accurate. However, \bl{in cases where the motion of a vehicle undergoes several} accelerations and decelerations, there may be a distortions in the 3D map due to
errors in motion estimation. Therefore, it is \bl{preferable to avoid making} assumptions about
the accuracy of the odometry to use the proposed method in all situations.
Therefore, if the uncertainty of the odometry and the SLAM framework are
included in the calibration framework, these two problems can be \bl{simultaneously} solved together
through the optimization process.


\bibliographystyle{IEEEtranN}
\bibliography{string-short,references}

\begin{thebibliography}{24}
\providecommand{\natexlab}[1]{#1}
\providecommand{\url}[1]{#1}
\csname url@samestyle\endcsname
\providecommand{\newblock}{\relax}
\providecommand{\bibinfo}[2]{#2}
\providecommand{\BIBentrySTDinterwordspacing}{\spaceskip=0pt\relax}
\providecommand{\BIBentryALTinterwordstretchfactor}{4}
\providecommand{\BIBentryALTinterwordspacing}{\spaceskip=\fontdimen2\font plus
\BIBentryALTinterwordstretchfactor\fontdimen3\font minus
  \fontdimen4\font\relax}
\providecommand{\BIBforeignlanguage}[2]{{%
\expandafter\ifx\csname l@#1\endcsname\relax
\typeout{** WARNING: IEEEtranN.bst: No hyphenation pattern has been}%
\typeout{** loaded for the language `#1'. Using the pattern for}%
\typeout{** the default language instead.}%
\else
\language=\csname l@#1\endcsname
\fi
#2}}
\providecommand{\BIBdecl}{\relax}
\BIBdecl

\bibitem[Shin et~al.(2018)Shin, Park, and Kim]{shin2018direct}
Y.-S. Shin, Y.~S. Park, and A.~Kim, ``Direct visual {SLAM} using sparse depth
  for camera-lidar system,'' in \emph{Proc. {IEEE} Intl. Conf. on Robot. and
  Automat.}, 2018, pp. 1--8.

\bibitem[Zhang and Singh(2015)]{zhang2015visual}
J.~Zhang and S.~Singh, ``Visual-lidar odometry and mapping: Low-drift, robust,
  and fast,'' in \emph{Proc. {IEEE} Intl. Conf. on Robot. and Automat.}, 2015,
  pp. 2174--2181.

\bibitem[Han et~al.(2017)Han, Wang, Lu, and Zhao]{han2017road}
X.~Han, H.~Wang, J.~Lu, and C.~Zhao, ``Road detection based on the fusion of
  lidar and image data,'' \emph{Intl. J. of Robot. Research}, vol.~14, no.~6,
  pp. 1--10, 2017.

\bibitem[Yousef et~al.(2017)Yousef, Mohd, Al-Widyan, and
  Hayajneh]{ahmad2017extrinsic}
K.~M.~A. Yousef, B.~J. Mohd, K.~Al-Widyan, and T.~Hayajneh, ``Extrinsic
  calibration of camera and {2D} laser sensors without overlap,'' \emph{{IEEE}
  Sensors J.}, vol.~17, no.~10, pp. 1--24, 2017.

\bibitem[Zhou(2014)]{zhou2014new}
L.~Zhou, ``A new minimal solution for the extrinsic calibration of a {2D} lidar
  and a camera using three plane-line correspondences,'' \emph{{IEEE} Sensors
  J.}, vol.~14, pp. 442--454, 2014.

\bibitem[Krishnan and Saripalli(2017)]{krishnan2017cross}
A.~K. Krishnan and S.~Saripalli, ``Cross-calibration of {RGB} and thermal
  cameras with a lidar for rgb-depth-thermal mapping,'' \emph{Unmanned Sys.},
  vol.~5, no.~02, pp. 59--78, 2017.

\bibitem[Per{\v{s}}i{\'c} et~al.(2017)Per{\v{s}}i{\'c}, Markovi{\'c}, and
  Petrovi{\'c}]{pervsic2017extrinsic}
J.~Per{\v{s}}i{\'c}, I.~Markovi{\'c}, and I.~Petrovi{\'c}, ``Extrinsic {6DoF}
  calibration of {3D LiDAR} and radar,'' in \emph{European Conf. on Mobile
  Robots}, 2017, pp. 1--6.

\bibitem[Alismail et~al.(2012)Alismail, Baker, and
  Browning]{alismail2012automatic}
H.~Alismail, L.~D. Baker, and B.~Browning, ``Automatic calibration of a range
  sensor and camera system,'' in \emph{Proc. {IEEE} Intl. Conf. on 3D Imaging,
  Modeling, Processing, Visualization and Transmission}, 2012, pp. 286--292.

\bibitem[Taylor and Nieto(2013)]{taylor2013automatic}
Z.~Taylor and J.~Nieto, ``Automatic calibration of {LiDAR} and camera images
  using normalized mutual information,'' in \emph{Proc. {IEEE} Intl. Conf. on
  Robot. and Automat.}, 2013.

\bibitem[Pandey et~al.(2015)Pandey, McBride, Savarese, and
  Eustice]{pandey2015automatic}
G.~Pandey, J.~R. McBride, S.~Savarese, and R.~M. Eustice, ``Automatic extrinsic
  calibration of vision and lidar by maximizing mutual information,'' \emph{J.
  of Field Robot.}, vol.~32, no.~5, pp. 696--722, 2015.

\bibitem[Iyer et~al.(2018)Iyer, Ram, Murthy, and Krishna]{iyer2018calibnet}
G.~Iyer, R.~K. Ram, J.~K. Murthy, and K.~M. Krishna, ``Calibnet: Geometrically
  supervised extrinsic calibration using {3D} spatial transformer networks,''
  in \emph{Proc. {IEEE}/{RSJ} Intl. Conf. on Intell. Robots and Sys.}, 2018,
  pp. 1110--1117.

\bibitem[Zhao et~al.(2007)Zhao, Chen, and Shibasaki]{zhao2007efficient}
H.~Zhao, Y.~Chen, and R.~Shibasaki, ``An efficient extrinsic calibration of a
  multiple laser scanners and cameras' sensor system on a mobile platform,'' in
  \emph{Proc. {IEEE} Intell. Vehicle Symposium}, 2007, pp. 422--427.

\bibitem[Tamas and Kato(2013)]{tamas2013targetless}
L.~Tamas and Z.~Kato, ``Targetless calibration of a lidar-perspective camera
  pair,'' in \emph{Proc. of the IEEE Int. Conf. on Comp. Vision Workshops},
  2013, pp. 668--675.

\bibitem[Scott et~al.(2015)Scott, Morye, Pini{\'e}s, Paz, Posner, and
  Newman]{scott2015exploiting}
T.~Scott, A.~A. Morye, P.~Pini{\'e}s, L.~M. Paz, I.~Posner, and P.~Newman,
  ``Exploiting known unknowns: Scene induced cross-calibration of lidar-stereo
  systems,'' in \emph{Proc. {IEEE}/{RSJ} Intl. Conf. on Intell. Robots and
  Sys.}, 2015, pp. 3647--3653.

\bibitem[Scott et~al.(2016)Scott, Morye, Pini{\'e}s, Paz, Posner, and
  Newman]{scott2016choosing}
------, ``Choosing a time and place for calibration of lidar-camera systems,''
  in \emph{Proc. {IEEE} Intl. Conf. on Robot. and Automat.}, 2016, pp.
  4349--4356.

\bibitem[Napier et~al.(2013)Napier, Corke, and Newman]{napier2013cross}
A.~Napier, P.~Corke, and P.~Newman, ``Cross-calibration of push-broom {2D}
  lidars and cameras in natural scenes,'' in \emph{Proc. {IEEE} Intl. Conf. on
  Robot. and Automat.}, 2013, pp. 3679--3684.

\bibitem[Taylor and Nieto(2016)]{taylor2016motion}
Z.~Taylor and J.~Nieto, ``Motion-based calibration of multimodal sensor
  extrinsics and timing offset estimation,'' \emph{{IEEE} Trans. Robot.},
  vol.~32, no.~5, pp. 1215--1229, 2016.

\bibitem[Andreff et~al.(2001)Andreff, Horaud, and Espiau]{andreff2001robot}
N.~Andreff, R.~Horaud, and B.~Espiau, ``Robot hand-eye calibration using
  structure-from-motion,'' \emph{Intl. J. of Robot. Research}, vol.~20, no.~3,
  pp. 228--248, 2001.

\bibitem[Heller et~al.(2011)Heller, Havlena, Sugimoto, and
  Pajdla]{heller2011structure}
J.~Heller, M.~Havlena, A.~Sugimoto, and T.~Pajdla, ``Structure-from-motion
  based hand-eye calibration using l$\infty$ minimization,'' in \emph{Proc.
  {IEEE} Conf. on Comput. Vision and Pattern Recog.}, 2011, pp. 3497--3503.

\bibitem[Jeong et~al.(2018)Jeong, Cho, Shin, Roh, and Kim]{jjeong2018icra}
J.~Jeong, Y.~Cho, Y.-S. Shin, H.~Roh, and A.~Kim, ``Complex urban {LiDAR} data
  set,'' in \emph{Proc. {IEEE} Intl. Conf. on Robot. and Automat.}, Brisbane,
  May. 2018, pp. 6344--6351.

\bibitem[Labayrade et~al.(2002)Labayrade, Aubert, and Tarel]{labayrade2002real}
R.~Labayrade, D.~Aubert, and J.-P. Tarel, ``Real time obstacle detection in
  stereovision on non flat road geometry through" v-disparity"
  representation,'' in \emph{Proc. {IEEE} Intell. Vehicle Symposium},
  vol.~2.\hskip 1em plus 0.5em minus 0.4em\relax IEEE, 2002, pp. 646--651.

\bibitem[Gioi et~al.(2010)Gioi, Jakubowicz, Morel, and Randall]{von2010lsd}
R.~G.~V. Gioi, J.~Jakubowicz, J.-M. Morel, and G.~Randall, ``{LSD}: A fast line
  segment detector with a false detection control,'' \emph{{IEEE} Trans.
  Pattern Analysis and Machine Intell.}, vol.~32, no.~4, pp. 722--732, 2010.

\bibitem[Rabbani et~al.(2006)Rabbani, Heuvel, and
  Vosselmann]{rabbani2006segmentation}
T.~Rabbani, F.~V.~D. Heuvel, and G.~Vosselmann, ``Segmentation of point clouds
  using smoothness constraint,'' \emph{Intl. archives of photogrammetry, remote
  sensing and spatial info. sci.}, vol.~36, no.~5, pp. 248--253, 2006.

\bibitem[Canny(1986)]{canny1986computational}
J.~Canny, ``A computational approach to edge detection,'' \emph{{IEEE} Trans.
  Pattern Analysis and Machine Intell.}, no.~6, pp. 679--698, 1986.

\end{thebibliography}

\end{document}